\pdfoutput=1

\documentclass[11pt]{article}

\usepackage[]{acl}

\usepackage{times}
\usepackage{latexsym}
\usepackage{booktabs}
\usepackage{multirow}
\usepackage{graphicx}
\usepackage{makecell}
\usepackage{bm}
\usepackage{amssymb}
\usepackage{amsmath}
\usepackage{arydshln}
\newcounter{paranum}
\newcommand{\paranum}{\refstepcounter{paranum}}
\newcommand{\labelpara}[1]{\paranum\label{#1}}

\usepackage[T1]{fontenc}

\usepackage[utf8]{inputenc}

\usepackage{microtype}

\title{Self-Training with Pseudo-Label Scorer for Aspect Sentiment Quad Prediction}

\author{Yice Zhang$^{1,3}$, Jie Zeng$^{1}$\thanks{\quad The three authors contribute equally to this work.}\ , Weiming Hu$^{1\ast}$, Ziyi Wang$^{1\ast}$, \\
\bf Shiwei Chen$^{1,2}$, and Ruifeng Xu$^{1,2,3}$\thanks{\quad Corresponding Authors}\\
 $^{1}$ Harbin Institute of Technology, Shenzhen, China \\
 $^{2}$ Peng Cheng Laboratory, Shenzhen, China \\
 $^{3}$ Guangdong Provincial Key Laboratory of Novel Security Intelligence Technologies \\
\texttt{zhangyc\_hit@163.com,}\texttt{xuruifeng@hit.edu.cn} \\
}

\begin{document}
\maketitle
\begin{abstract}

Aspect Sentiment Quad Prediction (ASQP)
aims to predict all quads (aspect term, aspect category, opinion term, sentiment polarity) for a given review, which is the most representative and challenging task in aspect-based sentiment analysis.
A key challenge in the ASQP task is the scarcity of labeled data, which limits the performance of existing methods.
To tackle this issue, we propose a self-training framework with a pseudo-label scorer, wherein a scorer assesses the match between reviews and their pseudo-labels, aiming to filter out mismatches and thereby enhance the effectiveness of self-training. 
We highlight two critical aspects to ensure the scorer's effectiveness and reliability: the quality of the training dataset and its model architecture. 
To this end, we create a human-annotated comparison dataset and train a generative model on it using ranking-based objectives.
Extensive experiments on public ASQP datasets reveal that using our scorer can greatly and consistently improve the effectiveness of self-training.
Moreover, we explore the possibility of replacing humans with large language models for comparison dataset annotation, and experiments demonstrate its feasibility.\footnote{We release our code and data at \url{https://github.com/HITSZ-HLT/ST-w-Scorer-ABSA}.}
\end{abstract}

\section{Introduction}

Aspect-Based Sentiment Analysis (ABSA) aims to recognize aspect-level opinions and sentiments from user-generated content \cite{pontiki-etal-2014-semeval}.
This problem has consistently attracted interest
owing to its proficiency in distilling and summarizing fine-grained opinions from vast data \cite{hai2019survey,Nazir2022survey,zhang2023survey}.
The most representative and challenging task in ABSA is Aspect Sentiment Quad Prediction (ASQP) \cite{cai-etal-2021-aspect,zhang-etal-2021-aspect-sentiment}. 
This task formulates aspect-level opinions and sentiments as quadruples, each consisting of an aspect term, aspect category, opinion term, and sentiment polarity.
For example, given a review ``\textit{the food is great and reasonably priced},'' the output of ASQP would be \{(\textit{food}, food\_quality, \textit{great}, positive), (\textit{food}, food \_prices, \textit{reasonably priced}, positive)\}.

\begin{figure}
\centering
\includegraphics[width=1\linewidth]{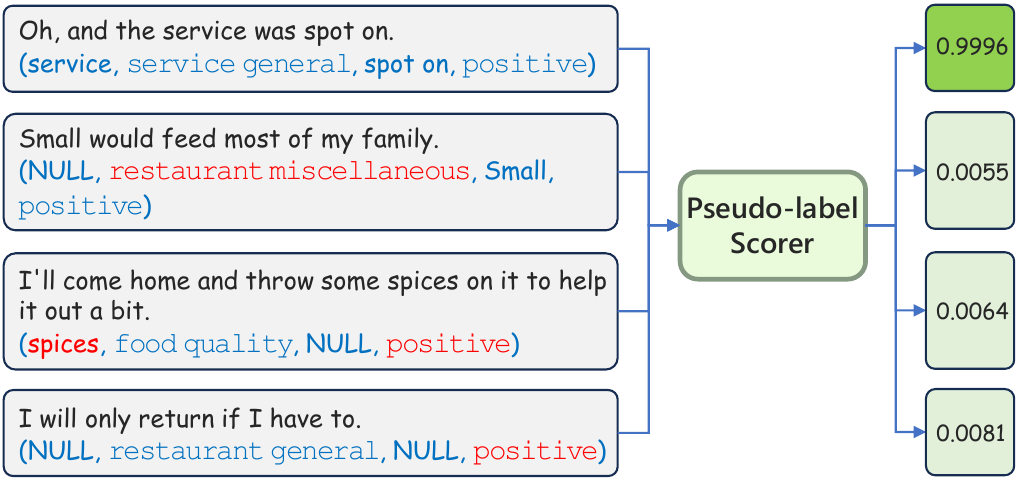}
\caption{Illustration of our pseudo-label scorer.}
\label{fig:scorer}
\end{figure}

As a fine-grained problem, ABSA faces the challenge of insufficient labeled data, which is particularly severe in the ASQP task. 
This issue limits the performance of existing models. 
Many efforts explore data augmentation methods to alleviate this issue.
They synthesize new samples by modifying existing ones \citep{li-etal-2020-conditional,hsu-etal-2021-semantics}, applying self-training techniques \citep{wang-etal-2021-progressive}, or utilizing generative methods \citep{yu-etal-2023-cross,deng-etal-2023-bidirectional,zhang-etal-2023-target-source}.
However, a significant limitation of these methods is that the synthetic samples often inevitably contain mismatches between sentences and labels, which can adversely affect model learning.

To reduce such mismatches, this paper introduces a pseudo-label scorer for data augmentation. 
As illustrated in Figure \ref{fig:scorer}, the scorer assesses the degree of match between the review and its pseudo-label.
If we have a sufficiently robust scorer, we can filter out all mismatched samples, thereby significantly enhancing the effectiveness of data augmentation.
We propose that the effectiveness and reliability of this scorer hinge on two critical aspects: (1) the quality of the training dataset and (2) its architecture along with the training objective.
We discuss these two aspects below.

For the first aspect, previous works typically produce negative labels by modifying real labels using heuristic rules \cite{wang-etal-2021-progressive, mao-etal-2022-seq2path}. However, such negative labels are usually simplistic and patterned, limiting the scorer's learning.
To overcome this limitation, we create a human-annotated comparison dataset. 
Specifically, we train an ASQP model with existing labeled data, use it to infer several pseudo-labels for unlabeled data, and then have human annotators choose the most appropriate pseudo-labels. 
The labels chosen by annotators are designated as positive labels, while the rest as negative labels. 
Our dataset, in contrast to the rule-based datasets,  is more challenging and better aligned with human judgment.

For the second aspect, previous works formalize label-scoring as a question-answering problem \cite{wang-etal-2021-progressive} or embed the discriminative matching token into the label \cite{mao-etal-2022-seq2path}. 
However, our findings suggest that these methods underperform in complex tasks like ASQP, 
due to their limited capacity to model the interactions between reviews and pseudo-labels.
Recent works in preference optimization reveal that the language model itself can serve as a scorer \citep{rafailov2023direct,yuan2023rrhf}. This motivates us to use the conditional likelihoods that a generative model assigns to a pseudo-label as the measure of its quality. 
Compared with the previous methods, this approach enables the scorer to examine the plausibility of a pseudo-label in a token-by-token fashion, thus offering a more comprehensive and effective scoring. 
We then fine-tune this scorer on our comparison dataset using ranking-based objectives.

Upon developing this pseudo-label scorer, we apply it in a data augmentation framework, specifically opting for the self-training framework due to its simplicity.
We conduct extensive experiments on public ASQP datasets to examine its effectiveness and further investigate the following questions:
(1) how does the pseudo-label scorer perform using our comparison data and model architecture?;
(2) is it feasible to replace humans with large language models to annotate the comparison data?; and
(3) how to utilize the scorer to filter out low-quality samples?
Furthermore, inspired by \citet{ma2023large}, we extend the application of this scorer, employing it as a reranker for multiple candidate labels, and assess its impact and effectiveness.

Our contributions are summarized as follows:
\begin{enumerate}
    \item[(1)] To the best of our knowledge, we are the first to apply a pseudo-label scorer to data augmentation in the ASQP task.
    \item[(2)] We investigate how to enhance the scorer's effectiveness and reliability from both dataset and model architecture perspectives.
    \item[(3)] We empirically demonstrate that the proposed pseudo-label scorer can significantly and consistently enhance the performance of existing models.
\end{enumerate}

\section{Background}

\textbf{Task Evolution.}
Aspect-Based Sentiment Analysis (ABSA) is a fine-grained sentiment analysis problem, aiming at recognizing user opinions and sentiments towards specific aspects \cite{zhang2023survey}. 
Within ABSA, aspects are generally defined as aspect categories or aspect terms. Aspect categories are predefined categories like food\_quality and service\_general in restaurant reviews, or laptop\_portability and display\_quality in laptop reviews. 
Aspect terms are explicit mentions of these aspect categories in the text.
Based on this, \citet{pontiki-etal-2014-semeval} formally define four sub-tasks: aspect category detection, aspect category sentiment classification, aspect term extraction, and aspect term sentiment classification. These tasks sequentially identify aspects and determine their corresponding sentiments.

Opinion term refers to words or phrases that express subjective sentiments. Sentiment expressions towards aspects often rely on these opinion terms, making them vital clues for recognizing aspects and their sentiment polarities.
Consequently, many researchers focus on aspect and opinion co-extraction \cite{wang-etal-2016-recursive,li-lam-2017-deep,fan-etal-2019-target,chen-etal-2020-synchronous,zhao-etal-2020-spanmlt}.
Building on this, \citet{peng2020aste} propose the Aspect Sentiment Triplet Extraction (ASTE) task, conceptualizing an aspect-level sentiment as a triplet consisting of an aspect term, opinion term, and sentiment polarity. 
Following ASTE, \citet{cai-etal-2021-aspect,zhang-etal-2021-aspect-sentiment} extend the triplet by incorporating the aspect category, evolving it into Aspect Sentiment Quad Prediction (ASQP). ASQP is currently the most representative and challenging task within ABSA.

\vspace{5pt}
\noindent
\textbf{ABSA Methods.}
Early efforts in ABSA primarily focus on identifying one or two sentiment elements. They mainly design specific model structures to establish interactions between sentiment elements and their contexts, as well as among different sentiment elements \cite{wang-etal-2016-recursive,li-lam-2017-deep,ma2017ian,chen-etal-2017-recurrent,xu-etal-2018-double,xue-li-2018-aspect,li-etal-2018-transformation,fan-etal-2019-target,lcf2019}.
Recent efforts have shifted towards compound tasks like ASTE and ASQP, proposing various end-to-end methods. These includes machine reading comprehension-based methods \cite{bmrc,dual-mrc}, table-filling methods \cite{wu-etal-2020-grid,chen-etal-2022-enhanced,zhang-etal-2022-boundary}, span-based methods \cite{xu-etal-2021-learning,LI2022108366,cai-etal-2021-aspect}, and generative methods \cite{yan-etal-2021-unified,zhang-etal-2021-towards-generative,zhang-etal-2021-aspect-sentiment,mao-etal-2022-seq2path,gao-etal-2022-lego,hu-etal-2023-improving,hu-etal-2023-uncertainty,gou-etal-2023-mvp}.
Among these, generative methods have emerged as mainstream due to their universality and capacity to exploit rich label semantics.

\vspace{5pt}
\noindent
\textbf{Generative Aspect-Based Sentiment Analysis}, abbreviated as GAS, is a unified generative framework proposed by \citet{zhang-etal-2021-towards-generative}. The core idea of this framework is to transform sentiment elements into a label sequence and then use a \textsc{seq2seq} model to learn the dependencies between the input text and the label sequence. For the ASQP task, we can convert it into a label sequence using the following template:
\begin{equation}
    {\rm seq}_{\rm label} = c_1\ |\ s_1\ |\ a_1\ |\ o_1\ ;\ \cdots\ ;\ c_n\ |\ s_n\ |\ a_n\ |\ o_n,\nonumber
\end{equation}
where $c_i$ denotes the aspect category, $s_i$ denotes the sentiment polarity, $a_i$ denotes the aspect term, and $o_i$ denotes the opinion term.

\section{Comparison Dataset}

We need to construct a comparison dataset to facilitate the training and evaluation of the pseudo-label scorer. This dataset comprises samples each containing a review sentence accompanied by several pseudo-labels, where one is the positive label and the others are negative labels.
We train the scorer by requiring it to assign high scores to positive labels and low scores to negative labels.

Previous works typically produce negative labels using heuristic rules \cite{wang-etal-2021-progressive,mao-etal-2022-seq2path}. 
They randomly modify elements in the existing positive labels, such as altering boundaries or conducting substitutions.
Such negative labels are patterned and easily distinguishable, limiting the learning potential of the scorer.
Therefore, this paper employs human annotators to construct this comparison dataset.

\subsection{Data Preparation}

Aligned with existing ASQP datasets \cite{cai-etal-2021-aspect,zhang-etal-2021-aspect-sentiment}, we collect reviews from two domains: \texttt{Restaurant} and \texttt{Laptop}.
The restaurant reviews are from the Yelp Dataset\footnote{\url{https://www.yelp.com/dataset}}, and the laptop reviews are from the Amazon Laptop Dataset\footnote{\url{https://nijianmo.github.io/amazon/index.html}} \cite{ni-etal-2019-justifying}.
We segment these reviews into individual sentences.
Next, we employ the existing labeled dataset to train an ASQP model \cite{zhang-etal-2021-towards-generative} and then utilize this model to generate four pseudo-labels for each review sentence via beam search.

\subsection{Annotation Process}

For a review sentence and its four pseudo-labels, annotators are presented with six options. The first four options correspond to the four pseudo-labels, the fifth option indicates that none of the pseudo-labels are appropriate, and the sixth option suggests that the review sentence does not express any sentiment or the expressed sentiment is difficult to infer. When annotators choose the fifth option, they are required to write an alternative label.

The annotation process is organized into multiple batches, each containing about 200 samples. To ensure accuracy, every sample is independently annotated by three different annotators. In cases of discrepancy among their annotations, a fourth annotator steps in to resolve the inconsistency. Furthermore, at the conclusion of each batch, the four annotators meet to discuss and reconcile any disagreements.
Each annotator is provided with annotation guidelines\footnote{\url{https://alt.qcri.org/semeval2014/task4/data/uploads/semeval14_absa_annotationguidelines.pdf} and
\url{https://alt.qcri.org/semeval2016/task5/data/uploads/absa2016_annotationguidelines.pdf}.} and existing labeled ASQP datasets. In instances where conflicts arise between the two, we prioritize adherence to the guidelines.

\vspace{5pt}
\noindent\textbf{AI Annotation.}
Choosing the most appropriate label, while much simpler than annotating an ASQP label from scratch, remains a laborious task. Therefore, we explore the feasibility of using ChatGPT as a substitute for human annotators.
To ensure the quality of AI annotations, we carefully craft prompts for each ASQP dataset. Additionally, we incorporate three strategies to enhance the annotation process: self-consistency, self-assessment, and rationale augmentation. 
Details of AI annotation can be found in Appendix \ref{sec:ai-annotation}.

\begin{table}[t]
\centering
\fontsize{8pt}{0.8\baselineskip}\selectfont
\setlength\tabcolsep{2pt}
\begin{tabular}{l cc c cc cc} 
\toprule
Datasets & P1 & P2 & P3 & P4 & P5 & P6 & Total \\
\midrule
\texttt{ACOS-Laptop-Comp}    & 853 & 149 & 88 & 44 & 167 & 204 & 1505\\
\texttt{ACOS-Rest-Comp}      & 
894 & 109 & 49 & 23 & 122 & 1003 & 2200 \\
\hdashline[2pt/4pt]
\texttt{ACOS-Laptop-Comp-AI} & 1744 & 410 & 213 & 122 & 0 & 259 & 2748\\
\texttt{ACOS-Rest-Comp-AI}   & 
1796 & 276 & 110 & 60 & 0 & 1620 & 3862 \\
\texttt{ASQP-Rest15-Comp-AI} & 1585 & 369 & 157 & 94 & 0 & 1223 & 3428 \\
\texttt{ASQP-Rest16-Comp-AI} & 1560 & 431 & 114 & 76 & 0 & 1337 & 3518 \\
\bottomrule
\end{tabular}
\caption{
Statistics of the comparison datasets. P1-P6 correspond to the number of samples for options 1 to 6, respectively.
}
\label{tab:comp-stat}
\end{table}

\subsection{Statistics}
We construct two human-annotated and four AI-annotated comparison datasets. 
Their basic statistical information is presented in Table \ref{tab:comp-stat}.
In the training phase of the scorer, we exclude samples corresponding to option 6 and reserve a portion of the data as the development set for hyperparameter tuning and model selection. Specifically, for the \texttt{Restaurant} datasets, we set aside 200 samples, and for the \texttt{Laptop} datasets, we allocate 300 samples. Consequently, this leaves around 1,000 training samples in the human-annotated datasets and approximately 2,000 training samples in the AI-annotated datasets.

\section{Our Approach}
\subsection{Pseudo-label Scorer}

The objective of the pseudo-label scorer is to score the match between a review and a pseudo-label. Previous efforts formalize this scoring task as a question-answering problem \cite{wang-etal-2021-progressive} or embed the discriminative matching token into the label \cite{mao-etal-2022-seq2path}. However, these methods struggle to effectively capture the interaction between reviews and pseudo-labels. Inspired by recent works in preference optimization \cite{rafailov2023direct,yuan2023rrhf,song2023preference}, we utilize a generative model as the scorer. Given a review sentence $x$ and a pseudo label $y$, their matching score is quantified by the conditional probability assigned by the generative model:
\begin{equation}
s(x,y)\propto p(y|x)=\prod_t p(y_t|y_{<t},x).
\end{equation}
Compared to previous methods, this approach integrates the likelihood of each token in the pseudo-label to derive its overall score, thereby providing a comprehensive and effective scoring.

\vspace{5pt}
\noindent\textbf{Training.} We optimize the pseudo-label scorer on the annotated comparison dataset with the ranking-based training objective.
Specifically, we design a simple listwise objective\footnote{
Besides the listwise objective, we also explore pointwise and pairwise objectives, which are presented in Appendix \ref{sec:ranking-objectives}.
} as follows:
\begin{align}
    {\cal L}_\textsc{list} &= - \log \frac{p(y_p|x)}{Z},\label{eq:list}\\
    Z &= p(y_p|x)+\sum_{y_n}p(y_n|x),
\end{align}
where $y_p$ denotes the positive label, $y_n$ denotes the negative label, and $Z$ is the normalization factor.

In addition to the comparison dataset, we also incorporate the original ASQP dataset to further enhance the training of the scorer. Labels from the original ASQP dataset are treated as additional positive labels and are combined with the positive labels from the comparison dataset. We additionally maximize the scores of these positive labels to enhance the scorer.
The combined loss function is formulated as follows:
\begin{align}
    {\cal L} =& {\cal L}_1 + \alpha {\cal L}_2,\\
    {\cal L}_1 =& {\mathbb{E}}_{(x,{\cal Y})\sim D_\textsc{comp}}{\cal L}_{\textsc{list}}(x,{\cal Y}),\\
    {\cal L}_2 =& {\mathbb{E}}_{(x,y_p)\sim D_{\textsc{comp}}\cup D_\textsc{asqp}}-\log p(y_p|x),
\end{align}
where $D_\textsc{comp}$ represents the comparison dataset, $D_\textsc{asqp}$ represents the original ASQP dataset, ${\cal Y}$ denotes the set of several pseudo-labels of the sentence $x$, and $\alpha$ is a hyperparameter.

\subsection{Self-Training with Data Filtering}

Self-training \cite{scudder1965}, a simple and classic semi-supervised technique, can be applied for data augmentation. It consists of three main steps: (1) training an initial model with the existing labeled dataset, (2) using this model to generate pseudo-labels for unlabeled data, and (3) finally incorporating these pseudo-labeled data into the labeled dataset.
However, this method inevitably introduces low-quality pseudo-labels, where the label does not accurately match the given review. To overcome this issue, we implement a two-stage filtering process that leverages both the initial model and the pseudo-label scorer.

\vspace{5pt}
\noindent\textbf{Confidence-based Filtering.}
We first use the confidence of the initial model in the pseudo-label as the measure of its quality. 
Thus, we filter out those samples with minimum confidence below a certain threshold. Formally, we retain samples $(x, y)$ satisfying
\begin{align}
    \left[\min_t p(y_t|y_{<t},x)\right] \ge \gamma_1,
\end{align}
where $\gamma_1$ is a hyper-parameter and is empirically set to 0.7.

\vspace{5pt}
\noindent\textbf{Scorer-based Filtering.}
Next, we use the pseudo-label scorer to evaluate the remaining samples. We observe that pseudo-labels with low scores are consistently of poor quality. Besides, while samples with high scores generally exhibit good label quality, their sentences tend to be overly simple, offering limited helpfulness for subsequent model training. Therefore, we retain only those samples whose scores fall between thresholds $\gamma_2$ and $\gamma_3$, which can be formulated as follows:
\begin{equation}
    \gamma_2 \le s(x,y) \le \gamma_3.
\end{equation}

\subsection{Pseudo-label Scorer as Reranker}

Reranking is originally a concept in information retrieval, referring to the process of rescoring and reranking preliminary candidate results. 
\citet{ma2023large} show that incorporating a reranking step can enhance performance in information extraction tasks.
In this paper, we claim that our pseudo-label scorer can serve as such a reranker.
Specifically, for a given review, we first utilize an ASQP model to generate four candidate labels via beam search and then select the best one from these candidates using our pseudo-label scorer.
The selected candidate is utilized as the final output.

\section{Experiments}
\subsection{Experiment Setup}

\begin{table}[t]
\centering
\fontsize{8.5pt}{0.8\baselineskip}\selectfont
\setlength\tabcolsep{3.95pt}
\begin{tabular}{l cc c cc c cc} 
\toprule
\multirow{2}*{{Datasets}} & \multicolumn{2}{c}{{Train}} && \multicolumn{2}{c}{{Dev}} && \multicolumn{2}{c}{{Test}} \\
\cmidrule(r){2-3} \cmidrule(r){5-6} \cmidrule(r){8-9}
& \#S & \#Q && \#S & \#Q  && \#S & \#Q\\
\midrule
\texttt{ACOS-Laptop} & 2934 & 4172 && 326 & 440 && 816 & 1161\\
\texttt{ACOS-Rest}   & 1530 & 2484 && 171 & 261 && 583 & 916\\
\texttt{ASQP-Rest15} & 834 & 1354 && 209 & 347 && 537 & 795 \\
\texttt{ASQP-Rest16} & 1264 & 1989 && 316 & 507 && 544 & 799\\
\bottomrule
\end{tabular}
\caption{
Statistics of four ASQP datasets \cite{cai-etal-2021-aspect,zhang-etal-2021-aspect-sentiment}. 
\#S and \#Q represent the number of sentences and quads. 
}
\label{tab:asqp-stat}
\end{table}

\textbf{Datasets.} 
We evaluate our approach on four public ASQP datasets. These datasets originate from the SemEval Challenges \cite{pontiki-etal-2015-semeval,pontiki-etal-2016-semeval} and Amazon platform during 2017 and 2018. The quad-level annotations are provided by \citet{cai-etal-2021-aspect} and \citet{zhang-etal-2021-aspect-sentiment}. Detailed statistics of these datasets are presented in Table \ref{tab:asqp-stat}.
Besides, to train the pseudo-label scorer, we construct several comparison datasets, the statistics of which can be found in Table \ref{tab:comp-stat}.

\vspace{5pt}
\noindent\textbf{Implementation Details.}
We utilize \texttt{T5-large} \cite{raffel2020} as the backbone for our pseudo-label scorer. During the training phase, we set both the batch size and the number of training epochs to 10. For other hyperparameters, including the learning rate and $\alpha$, we perform a simple hyperparameter search.
Once the scorer is trained, we apply it to score and rank the pseudo-labeled samples\footnote{
The source for the pseudo-labeled data is identical to that for the comparison data, 
originating from the Yelp Dataset and Amazon Laptop Dataset. But there is no overlap between them. 
The ASQP model used for pseudo-labeling is GAS \cite{zhang-etal-2021-towards-generative}.
}.
For datasets \texttt{ACOS-Rest}, \texttt{ASQP-Rest15}, and \texttt{ASQP-Rest16}, we retain samples with scores falling within the top 10\% to 40\%; for the \texttt{ACOS-Laptop} dataset, this range is set from 20\% to 50\%. From these retained samples, we randomly select 10,000 samples and merge them with the original labeled dataset to form the augmented dataset.
To reduce the impact of randomness, we run our approach five times and report the average results.

\vspace{5pt}
\noindent
\textbf{Baselines.}
To validate the effectiveness of the proposed approach, we integrate it into two typical ASQP methods: GAS \cite{zhang-etal-2021-towards-generative} and MUL \cite{hu-etal-2023-uncertainty}. 
We run these two methods on the augmented dataset and incorporate a reranking step during the inference phase to enhance the predictions.
We also benchmark our approach against a range of other methods, including \textsc{Extract-Classify} \cite{cai-etal-2021-aspect}, \textsc{Paraphrase} \cite{zhang-etal-2021-aspect-sentiment}, \textsc{Seq2Path} \cite{mao-etal-2022-seq2path}, \textsc{DLO/ILO} \cite{hu-etal-2022-improving-aspect}, \textsc{LEGO-ABSA} \cite{gao-etal-2022-lego}, \textsc{MvP} \cite{gou-etal-2023-mvp}, \textsc{GenDA} \cite{wang-etal-2023-generative}, and \textsc{ChatGPT} (few-shot) \cite{xu2023limits}.

\subsection{Analysis of Pesudo-label Scorer}

\begin{table}[t]
\centering
\fontsize{8.5pt}{0.8\baselineskip}\selectfont
\begin{tabular}{l cc c cc c} 
\toprule
Objectives & \texttt{ACOS-Laptop} & \texttt{ACOS-Rest} \\
\midrule
\citet{wang-etal-2021-progressive}  & 50.67 & 56.10 \\
\citet{mao-etal-2022-seq2path} & 64.34 & 72.00 \\
\textbf{Ours}  & \textbf{67.74} & \textbf{78.50} \\
\bottomrule
\end{tabular}
\caption{
Comparison results of the architecture for the pseudo-label scorer (accuracy, \%).
\citet{wang-etal-2021-progressive} formalize label-scoring as a question-answering problem.
\citet{mao-etal-2022-seq2path} append a discriminative matching token to the label.
}
\label{tab:scorer-model}
\end{table}

Given the importance of the pseudo-label scorer in our framework, we first undertake an analysis of it, focusing on two key aspects: its model architecture and the training dataset.

\vspace{5pt}
\noindent
\textbf{Model Architecture.}
We use the conditional likelihood the generative model assigns to a pseudo-label as its scoring metric. 
To examine the effectiveness of our approach, we conduct experiments on two human-annotated comparison datasets and benchmark our approach against previous methods \cite{wang-etal-2021-progressive,mao-etal-2022-seq2path}.
As Table \ref{tab:scorer-model} illustrates, previous methods, especially the question-answering method, perform poorly in the ASQP task. In contrast, our approach achieves a significant advantage, demonstrating its effectiveness.

\begin{table}[t]
\centering
\fontsize{8.5pt}{0.8\baselineskip}\selectfont
\begin{tabular}{l cc c cc c} 
\toprule
Annotation Schemes & \texttt{ACOS-Laptop} & \texttt{ACOS-Rest} \\
\midrule
\textsc{None}        & 60.53 & 74.10\\
\textsc{HumAnn-1234}   & 63.60 & 75.00\\
\textsc{HumAnn-12345}  & 64.60 & 76.60\\
\textsc{HumAnn-12345*} & 67.74 & 78.50\\
\textsc{AIAnn-1234*} & \textbf{68.67} & \textbf{79.60} \\
\bottomrule
\end{tabular}
\caption{
Experimental results of annotating the comparison dataset (accuracy, \%):
(1) \textsc{None} denotes the approach where neither human nor AI annotations are used, and the pseudo-label with the highest model confidence is selected as the positive label;
(2) \textsc{HumAnn-1234} represents the annotation scheme where human annotators choose the best pseudo-label out of four;
(3) \textsc{HumAnn-12345} extends \textsc{HumAnn-1234} by allowing human annotators to write an additional label when none of the four options are suitable;
(4) \textsc{AIAnn-1234} mirrors \textsc{HumAnn-1234}, but with ChatGPT replacing human annotators;
(5) methods with {*} indicate the training of the scorer using both the comparison dataset and the original ASQP dataset.
}
\label{tab:scorer-ann}
\end{table}

\begin{table}[t]
\centering
\fontsize{8pt}{0.8\baselineskip}\selectfont
\setlength\tabcolsep{3.8pt}
\begin{tabular}{l cc cc} 
\toprule
\multirow{2}*{{Datasets}} & \multicolumn{2}{c}{\texttt{w/ P6}} & \multicolumn{2}{c}{\texttt{w/o P6}} \\
\cmidrule(r){2-3} \cmidrule(r){4-5}
& \texttt{Kappa} & \texttt{Accu} & \texttt{Kappa} & \texttt{Accu} \\
\midrule
\texttt{ACOS-Laptop-Comp-AI}  & 62.71 & 79.20 & 65.84 & 86.30\\
\texttt{ACOS-Rest-Comp-AI}  & 67.44 & 80.90 & 47.12 & 87.15\\
\bottomrule
\end{tabular}
\caption{
Consistency between AI- and human-annotated comparison Data (\%). \texttt{P6} refers to samples with option 6 selected. We calculate the consistency both before and after removing these samples.
}
\label{tab:comp-consistency}
\end{table}

\vspace{5pt}
\noindent
\textbf{Comparison Dataset.}
We conduct experiments to compare different annotation schemes and list the results in Table \ref{tab:scorer-ann}. 
We have the following observations.
(1) Utilizing humans or AI to annotate the comparison data is crucial, as their performance is noticeably superior to that without annotation. Particularly, allowing human annotators to write a label when no option is suitable can notably enhance performance.
(2) Training the scorer with a combination of the comparison data and the original ASQP dataset is more effective than using the comparison data alone.
(3) AI-annotated comparison data can achieve even better results than human-annotated comparison data.

\begin{figure}[t]
\centering
\includegraphics[width=0.88\linewidth]{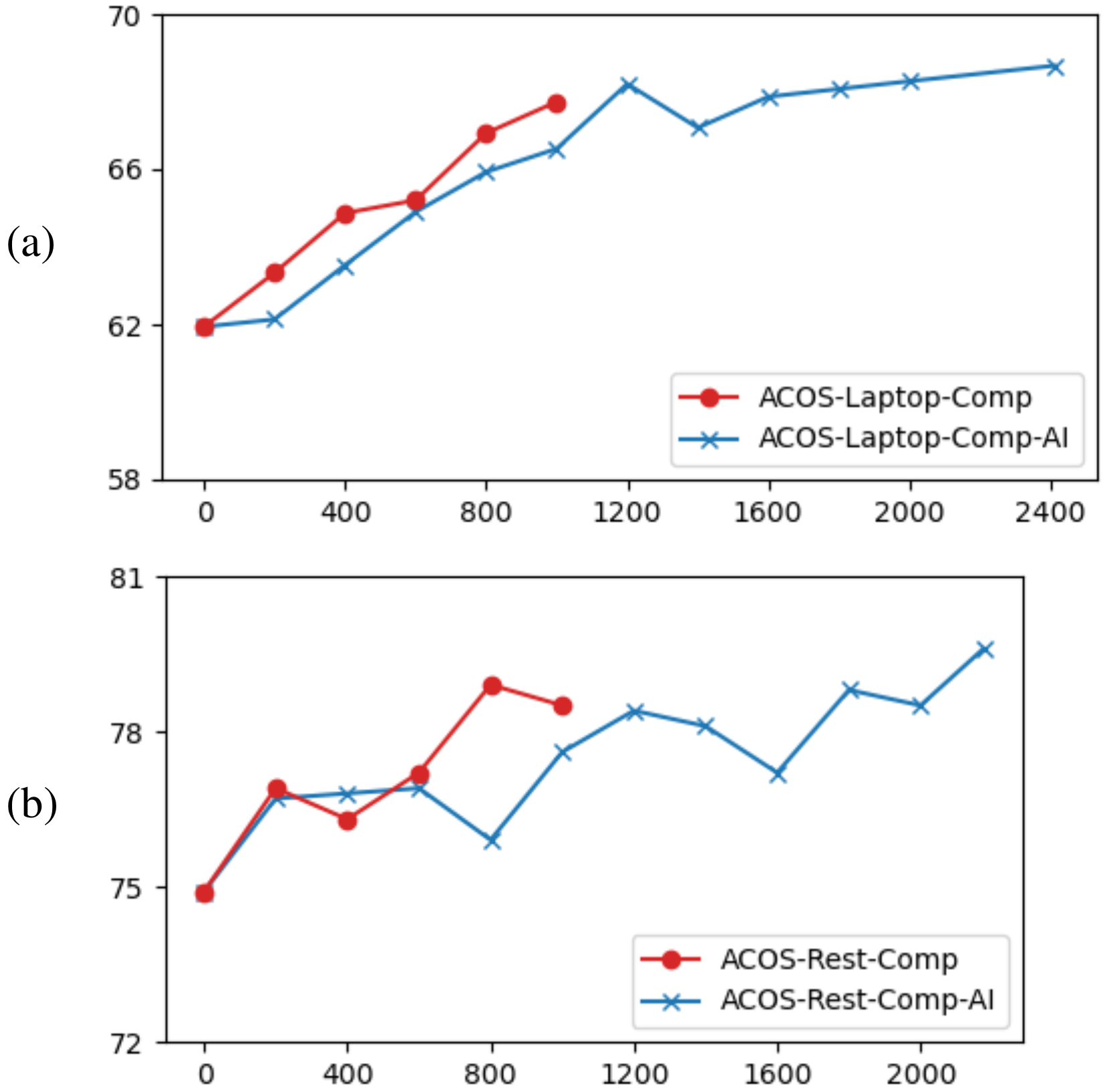}
\caption{
Performance trends of comparison data with increasing data quantity (accuracy, \%): (a) results on \texttt{ACOS-Laptop}; (b) results on \texttt{ACOS-Rest}.
}
\label{fig:scorer-data}
\end{figure}

\begin{table*}[t]
\centering
\fontsize{8.3pt}{0.8\baselineskip}\selectfont
\setlength\tabcolsep{2.8pt}
\begin{tabular}{l c ccc c ccc c ccc c ccc} 
\toprule
\multirow{2}*{{Methods}}  && \multicolumn{3}{c}{\texttt{ACOS-Laptop}} && \multicolumn{3}{c}{\texttt{ACOS-Rest}} && \multicolumn{3}{c}{\texttt{ASQP-Rest15}} && \multicolumn{3}{c}{\texttt{ASQP-Rest16}} \\
\cmidrule(r){3-5}  \cmidrule(r){7-9} \cmidrule(r){11-13} \cmidrule(r){15-17}
&& \texttt{Pre} & \texttt{Rec} & \texttt{F1} && \texttt{Pre} & \texttt{Rec} & \texttt{F1} && \texttt{Pre} & \texttt{Rec} & \texttt{F1} && \texttt{Pre} & \texttt{Rec} & \texttt{F1} \\
\midrule
\textsc{Extract-Classify} \cite{cai-etal-2021-aspect} && 45.56 & 29.28 & 35.80 && 38.54 & 52.96 & 44.61 && 35.64 & 37.25 & 36.42 && 38.40 & 50.93 & 43.77 \\
\textsc{Paraphrase} \cite{zhang-etal-2021-aspect-sentiment} && - & - & - && - & - & - && 46.16 & 47.72 & 46.93 && 56.63 & 59.30 & 57.93 \\
\textsc{Seq2Path} \cite{mao-etal-2022-seq2path} && - & - & 42.97 && - & - & 58.41 && - & - & -  && - & - & - \\
DLO \cite{hu-etal-2022-improving-aspect} && 43.40 & 43.80 & 43.60 && 60.02 & 59.84 & 59.18 && 47.08 & 49.33 & 48.18 && 57.92 & 61.80 & 59.79\\
ILO \cite{hu-etal-2022-improving-aspect} && 44.14 & 44.56 & 44.35 && 58.43 & 58.95 & 58.69 && 47.78 & 50.38 & 49.05 && 57.58 & 61.17 & 59.32\\
LEGO-ABSA \cite{gao-etal-2022-lego} && - & - & - && - & - & - && - & - & 45.80 && - & - & 57.70 \\
\textsc{MvP} \cite{gou-etal-2023-mvp} && - & - & 43.92 && - & - & 61.54 && - & - & 51.04 && - & - & 60.39\\
\textsc{GenDA} \cite{wang-etal-2023-generative} && - & - & - && - & - & - && 49.74 & 50.29 & 50.01 && 60.08 & 61.70 & 60.88\\
\textsc{ChatGPT} (few-shot) \cite{xu2023limits} && 21.72 & 27.65 & 24.33 & & 38.39 & 46.40 & 42.02 & & 29.66 & 37.86 & 33.26 & & 36.09 & 46.93 & 40.81\\
\midrule
GAS \cite{zhang-etal-2021-towards-generative} && 43.46 & 42.69 & 43.07 && 59.81 & 57.51 & 58.63 && 47.15 & 46.01 & 46.57 && 57.30 & 57.82 & 57.55 \\ 
~~+~ST && 44.35 & 42.75 & 43.54 && 59.95 & 58.98 & 59.46 && 46.86 & 47.65 & 47.25 && 57.95 & 58.75 & 58.35 \\
~~+~ST \& \textsc{C-Filter} && 45.14 & 44.00 & 44.56 && 60.57 & 58.82 & 59.67 && 49.04 & 47.77 & 48.40 & & 59.18 & 59.72 & 59.45 \\
~~+~ST \& \textsc{CS-Filter} && 46.23 & 44.41 & 45.30 && 63.41 & 60.00 & 61.66 && - & - & - && - & - & -\\ 
~~+~ST \& \textsc{CS-Filter} \& \textsc{ReRank} && \textbf{46.76} & \textbf{45.00} & \textbf{45.86} && \textbf{64.66} & \textbf{61.33} & \textbf{62.95}  &&  - & - & - && - & - & -\\ 
\hdashline[2pt/4pt]
~~+~ST \& \textsc{CS-Filter} (AI) && 46.44 & 44.01 & 45.19 && 62.69 & 60.24 & 61.44 && 50.92 & 49.86 & 50.38 && 60.87 & 61.30 & 61.08\\
~~+~ST \& \textsc{CS-Filter} \& \textsc{ReRank} (AI) && \textbf{47.00} & \textbf{45.05} & \textbf{46.01} && \textbf{63.74} & \textbf{61.25} & \textbf{62.47} && \textbf{51.59} & \textbf{51.90} & \textbf{51.74} && \textbf{62.55} & \textbf{64.31}	& \textbf{63.51}\\
\midrule
MUL \cite{hu-etal-2023-uncertainty} && 44.38 & 43.65 & 44.01 && 61.22 & 59.87 & 60.53 && 49.12 & 50.39 & 49.75 && 59.24 & 61.75 & 60.47 \\
\hdashline[2pt/4pt]
MUL (\textit{Our Reproduction}) && 42.79 & 41.95 & 42.45 && 61.22 & 59.80 & 60.50 && 48.28 & 49.74 & 48.99 && 58.42 & 60.68 & 59.52\\ 
~~+~ST  && 43.38 & 42.98 & 43.23 && 61.17 & 59.89 & 60.67 && 47.94 & 49.21 & 48.57 && 57.45 & 59.37 & 58.39 \\
~~+~ST \& \textsc{C-Filter}  && 44.59 & 43.67 & 44.13 && 62.11 & 60.20 & 61.14 && 48.74 & 49.06 & 48.90 && 59.44 & 61.07 & 60.25 \\ 
~~+~ST \& \textsc{CS-Filter} && 44.67 & 43.72 & 44.19 && 63.57 & 60.67 & 62.09 &&  - & - & - && - & - & - \\
~~+~ST \& \textsc{CS-Filter} \& \textsc{ReRank} && \textbf{46.88} & \textbf{44.74} & \textbf{45.78} && \textbf{66.18} & \textbf{61.75} & \textbf{63.89} &&  - & - & - && - & - & - \\
\hdashline[2pt/4pt] 
~~+~ST \& \textsc{CS-Filter} (AI) && 44.89 & 44.07 & 44.47 && 64.28 & 61.31 & 62.76 && 50.78 & 51.17 & 50.97 && 61.39 & 62.68 & 62.03 \\
~~+~ST \& \textsc{CS-Filter} \& \textsc{ReRank} (AI) && \textbf{47.05} & \textbf{45.32} & \textbf{46.17} && \textbf{65.43} & \textbf{61.92} & \textbf{63.63} && \textbf{51.94} & \textbf{52.00} &\textbf{51.97} && \textbf{63.46} & \textbf{64.31} & \textbf{63.88}\\
\bottomrule
\end{tabular}
\caption{
Experimental results on four ASQP datasets (\%).
\textsc{C-Filter} indicates the application of confidence-based filtering.
\textsc{CS-Filter} represents the integration of confidence-based and scorer-based filtering.
Methods marked with AI indicate that the pseudo-label scorer used is trained on AI-annotated comparison data.
}
\label{tab:main-result}
\end{table*}

We conduct a further analysis of AI Annotation. Table \ref{tab:comp-consistency} presents the consistency between AI- and human-annotated data. Although the consistency is not very high statistically,
considering the subjective nature of this task, the quality of AI annotation is acceptable. Additionally, a significant advantage of AI annotation lies in its cost-effectiveness relative to human annotation, enabling the efficient acquisition of a large amount of annotated data.

Figure \ref{fig:scorer-data} illustrates the performance trends of human- and AI-annotated data relative to their quantities. 
Although AI-annotated data exhibits lower performance at the same quantity, the scalability of AI annotation allows it to catch up and potentially exceed human-annotated data's performance when using more data.
For instance, more than 2,000 AI-annotated samples can equal or outperform 1,000 human-annotated samples.
Consequently, we can conclude that for the ASQP task, it is feasible to replace humans with AI to annotate the comparison data. 

\subsection{Analysis of Self-Training}

\textbf{Main Results.} 
We develop a self-training framework using the pseudo-label scorer, with the experimental results presented in Table \ref{tab:main-result}.
According to these results, our approach substantially and consistently improves the performance of existing ASQP methods \cite{zhang-etal-2021-towards-generative,hu-etal-2023-uncertainty}. Specifically, GAS achieves $F_1$-score improvements of 2.94\%, 4.32\%, 5.17\%, and 5.96\% across the four datasets, averaging at 4.60\%; MUL achieves $F_1$-score improvements of 3.72\%, 3.39\%, 2.98\%, and 4.36\% across these datasets, averaging at 3.61\%.
Upon integrating our approach, both GAS and MUL outperform previous methods.
These results demonstrate the effectiveness of our approach.

Furthermore, we have the following observations.
(1) The two-stage filtering process, namely \textsc{CS-Filter}, greatly enhances the effectiveness of self-training. In most datasets, it results in over 2\% improvement compared to self-training alone, highlighting the importance of data filtering in the self-training framework.
(2) Incorporating a rerank step can further improve performance, by around 1\%.
(3) Using AI-annotated data in downstream self-training can attain results comparable to those using human-annotated data. This further indicates the feasibility of replacing human annotators with AI for comparison data annotation.
(4) ChatGPT performs poorly on the ASQP task, suggesting that using it directly for this task does not fully leverage its capabilities. Conversely, using it for comparison data annotation effectively exploits its strengths.
(5) It can be noted that our filtering strategy offers relatively limited improvements on \texttt{ACOS-Laptop}. We attribute this to potential inconsistency between its ASQP annotations and our comparison annotations. A more detailed discussion is available in 
\hyperref[para:further3]{Further Analysis}.

\begin{figure}[t]
\centering
\includegraphics[width=1\linewidth]{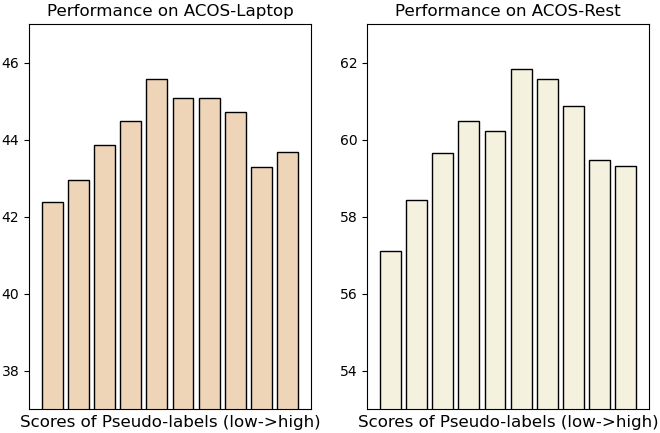}
\caption{Performance of GAS on the augmented dataset under different match scores ($F_1$-score, \%).}
\label{fig:match-score}
\end{figure}

\begin{figure}[t]
\centering
\includegraphics[width=0.89\linewidth]{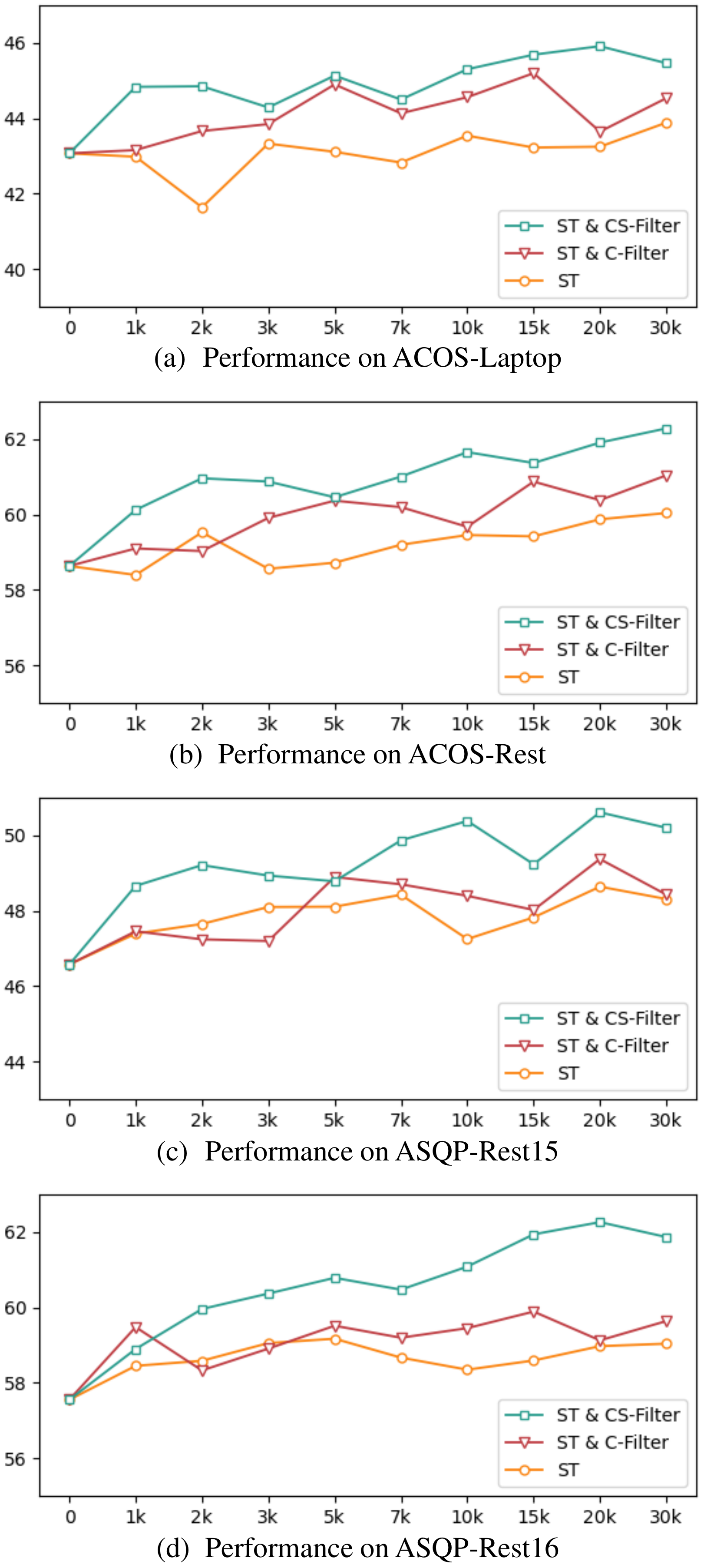}
\caption{Performance of GAS under different numbers of augmented samples ($F_1$-score, \%).}
\label{fig:data-quantity}
\end{figure}

\vspace{5pt}
\noindent
\textbf{Effect of Match Score.}
Our approach relies on the match scores output by the pseudo-label scorer for data filtering. 
We conduct experiments to examine the impact of these scores on self-training performance. Figure \ref{fig:match-score} illustrates that the performance incrementally increases with increasing match scores. Nevertheless, beyond a certain threshold, further increases in match scores lead to a decline in performance.
This phenomenon confirms our hypothesis that samples with too low scores suffer from poor label quality, adversely affecting model learning, and samples with too high scores tend to be overly simple, providing limited helpfulness for subsequent model training.

\vspace{5pt}
\noindent
\textbf{Effect of Data Quantity.} 
The quantity of pseudo-labeled samples is another important factor in the effectiveness of self-training. We conduct experiments to analyze its impact. As illustrated in Figure \ref{fig:data-quantity}, there is an overall upward trend in performance with increased data quantity. Notably, this trend is more stable and pronounced following two-stage filtering, underscoring the necessity of data filtering.
Furthermore, we notice a decrease in self-training performance when the number of augmented samples exceeds 20,000. This suggests that there is a limit to improving performance by simply increasing data quantity. Balancing diversity and label quality to enhance the effectiveness of self-training warrants further exploration in subsequent research.

\subsection{Further Analysis}

\begin{table}[t]
\centering
\fontsize{8pt}{0.8\baselineskip}\selectfont
\setlength\tabcolsep{2.5pt}
\begin{tabular}{l cccc c} 
\toprule
Methods & {Laptop} & {Rest} & {Rest15} & {Rest16} & Avg\\
\midrule
GAS    & 43.07 & 58.63 & 46.57 & 57.55 & -\\
\hdashline[2pt/4pt] 
+ ST \& \textsc{C-Filter} (1k)   & 43.16 & 59.10 & 47.45 & 59.48 & +0.84\\
+ ST \& \textsc{C-Filter} (2k)   & 43.66 & 59.03 & 47.24 & 58.33 & +0.61\\
\hdashline[2pt/4pt] 
+ ST \& \textsc{C-Filter} (10k)   & 44.56 & 59.67 & 48.40 & 59.45 & +1.57\\
+ ST \& \textsc{CS-Filter} (10k)  & 45.30 & 61.66 & 50.38 & 61.08 & +3.15\\
\hdashline[2pt/4pt] 
+ \textsc{CompData} (1k)         & 43.51 & 60.20 & - & - & -\\
+ \textsc{CompData (2k, AI)}    & 42.99 & 59.77 & 48.15 & 59.73 & +1.20\\
\bottomrule
\end{tabular}
\caption{
Experimental results of using comparison data as additional labeled data ($F_1$-scorer, \%).
The human- and AI-annotated comparison datasets contain about 1,000 and 2,000 samples, respectively.
}
\label{tab:comp-additional-data}
\end{table}

\labelpara{para:further1}
\textbf{Comparison Data as Additional Labeled Data.} 
One feasible approach for utilizing comparison data is to treat each sample along with its positive label as an additional labeled ASQP sample.
We analyze the effectiveness of this approach, and the results in Table \ref{tab:comp-additional-data} reveal:
(1) this approach can improve performance, with human-annotated comparison data outperforming AI-annotated data;
(2) under conditions of equal data volume, it is superior to self-training without data filtering, indicating that the quality of comparison data is better than that of pseudo-labeled data; and
(3) however, it falls significantly short of self-training with data filtering. 
These findings suggest that utilizing comparison data to train a pseudo-label scorer is more effective than simply treating it as additional labeled data.

\begin{table}[t]
\centering
\fontsize{8pt}{0.8\baselineskip}\selectfont
\setlength\tabcolsep{3.17pt}
\begin{tabular}{l cccc} 
\toprule
Methods & {Laptop} & {Rest} & {Rest15} & {Rest16}\\
\midrule
GAS    & 43.07 & 58.63 & 46.57 & 57.55\\
+ {Our Approach} & {45.86} & {62.95} & - & - \\
+ {Our Approach (AI)}  & {46.01} & {62.47} & {51.74} & {63.51}\\
\hdashline[2pt/4pt] 
\textsc{Pseudo-label Scorer}      & 43.93 & 60.66 & - & - \\
\textsc{Pseudo-label Scorer} (AI) & 44.06 & 60.00 & {51.59} & 61.49 \\
\bottomrule
\end{tabular}
\caption{
Experimental results of using pseudo-label scorer as the ASQP model ($F_1$-scorer, \%).
}
\label{tab:scorer-as-asqp-model}
\end{table}

\labelpara{para:further2}
\vspace{5pt}
\noindent
\textbf{Pseudo-label Scorer as the ASQP Model.} 
The pseudo-label scorer is architecturally a generative model and can potentially be used as an ASQP model. We evaluate this possibility and list the results in Table \ref{tab:scorer-as-asqp-model}. A surprising finding is that directly using the scorer to predict quads achieves good performance, though it generally falls short of using it for filtering and ranking. 
This suggests that, besides training the scorer, leveraging comparison data to enhance the ASQP model is a promising direction, deserving of in-depth investigation in future research.

\begin{table}[t]
\centering
\fontsize{8pt}{0.8\baselineskip}\selectfont
\begin{tabular}{c cccc} 
\toprule
 & {Laptop} & {Rest} & {Rest15} & {Rest16}\\
\midrule
$<0.1$ & 19.12\% & 3.92\%  & 3.24\%  & 1.50\% \\
$<0.3$ & 30.30\% & 7.32\%  & 5.52\%  & 2.37\% \\
$<0.5$ & 40.52\% & 10.46\% & 7.07\%  & 3.56\% \\
$<0.7$ & 52.56\% & 14.58\% & 9.59\%  & 5.46\% \\
$<0.9$ & 70.25\% & 25.88\% & 15.83\% & 9.43\% \\
\bottomrule
\end{tabular}
\caption{
Statistics of match scores in the ASQP training datasets.
}
\label{tab:match-score}
\end{table}

\begin{table}[t]
\centering
\fontsize{8pt}{0.8\baselineskip}\selectfont
\setlength\tabcolsep{4.7pt}
\begin{tabular}{c cccc c} 
\toprule
Ratio of Removal & {Laptop} & {Rest} & {Rest15} & {Rest16} & Avg \\
\midrule
0\% & 43.07 & 58.63 & 46.57 & 57.55 & -\\
\hdashline[2pt/4pt]
2\% & 43.38 & 59.44 & 47.20 & 58.27 & +0.62\\
4\% & 43.50 & \textbf{59.51} & \textbf{48.09} & 59.10 & \textbf{+1.10}\\
6\% & 43.89 & 59.24 & 46.78 & 58.25 & +0.59\\
8\% & 43.42 & 59.38 & 46.68 & 58.43 & +0.52\\
10\% & \textbf{44.25} & 59.09 & 46.57 & \textbf{59.20} & +0.82\\
\bottomrule
\end{tabular}
\caption{
Performance after removing samples with low match scores in the training set ($F_1$-scorer, \%).
}
\label{tab:removal}
\end{table}

\labelpara{para:further3}
\vspace{5pt}
\noindent
\textbf{Assessing Label Quality in ASQP Data.}
Beyond assessing pseudo-labeled data, our scorer can assess the quality of existing ASQP data. We conduct a statistical analysis of match scores for ASQP samples and list the results in Table \ref{tab:match-score}. 
Our analysis reveals relatively low match scores for the \texttt{ACOS-laptop} dataset, suggesting either poor annotation quality or low consistency with our comparison data. 
We manually review 100 samples with match scores low 0.1 and find that 73\% of the data contradicts the annotation guidelines, including 44\% with errors in aspect category annotations, 8\% in aspect or opinion term annotations, and 6\% in sentiment annotations.
Moreover, we experiment with removing samples with low match scores. The results presented in Table \ref{tab:removal} show that this removal not only preserves model performance but enhances it.
These findings indicate that our scorer is an effective tool for assessing label quality in existing datasets and that the removal of low-quality samples is advantageous.

\vspace{5pt}
\noindent
\textbf{Analysis of Reranking.}
We present the analysis of the reranking step in Appendix \ref{sec:analysis-of-ranking}.

\section{Conclusions}
In this paper, we introduce a pseudo-label scorer for the  Aspect Sentiment Quad Prediction (ASQP) task to reduce mismatches in data augmentation. We propose that the effectiveness and reliability of this scorer hinge on two critical aspects: the quality of the training dataset and its model architecture. 
To this end, we create both human- and AI-annotated comparison datasets and propose a scoring method based on a generative model. 
Upon developing this scorer, we apply it to data filtering in a self-training framework and further employ it as a reranker to enhance ASQP models. 
Detailed experiments and analysis demonstrate the effectiveness of our comparison datasets and the proposed architecture. Furthermore, experimental results on four public ASQP datasets reveal that our scorer significantly and consistently improves the performance of existing methods.

\newpage
\section*{Acknowledgements}
We thank the anonymous reviewers for their
valuable suggestions to improve the quality of this work.
We also thank Qianlong Wang for the enlightening discussions that significantly contributed to this work.
This work was partially supported by the National Natural Science Foundation of China 62176076, Natural Science Foundation of Guangdong 2023A1515012922, the Shenzhen Foundational Research Funding JCYJ20220818102415032, the Major Key Project of PCL2021A06, and Guangdong Provincial Key Laboratory of Novel Security Intelligence Technologies 2022B1212010005.

\section*{Limitations}

While our approach significantly enhances the effectiveness of data augmentation and improves the performance of existing ASQP models, it also suffers from the following limitations:
\begin{itemize}
 \item Data augmentation generally comprises two pivotal components: data synthesis and quality control. While this paper focuses primarily on the latter, the former is equally vital for the success of data augmentation. Given that models trained on limited labeled data may underperform in certain categories or contexts, targeted data synthesis can mitigate these issues. A comprehensive exploration of both data synthesis and quality control is essential for developing an effective and robust data augmentation framework.
\item The implementation of our approach necessitates manually annotated comparison data. Although we could use large language models to replace human annotators, crafting and refining prompts still demands meticulous human expertise and is notably time-intensive.
\end{itemize}
We argue that these limitations offer promising directions for future research.

\newpage
\bibliography{tacl2021,anthology}

\appendix
\section{Details of AI Annotation}
\label{sec:ai-annotation}

\begin{figure}[t]
\centering
\includegraphics[width=1.\linewidth]{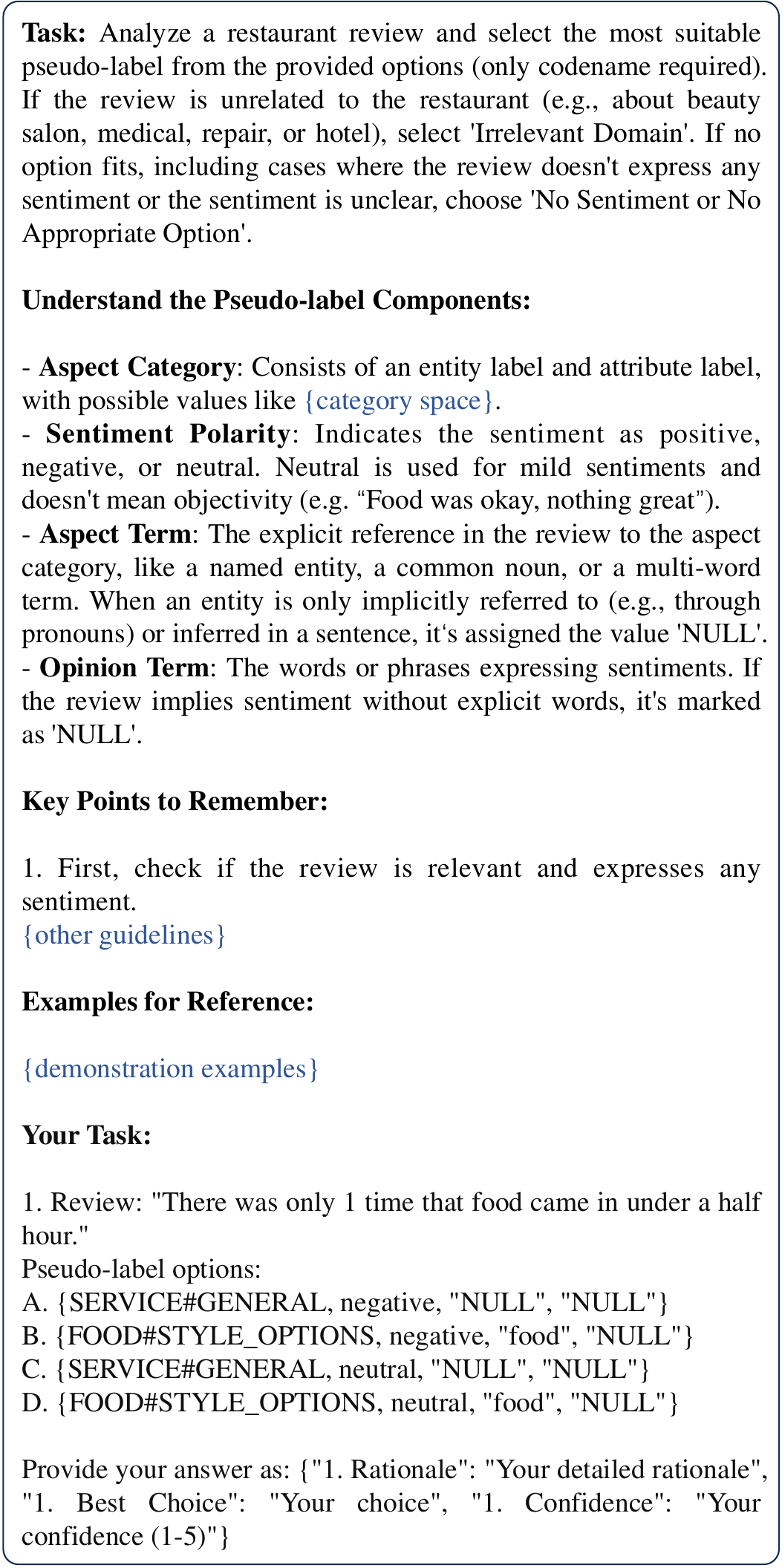}
\caption{Prompt for AI Annotation in \texttt{ACOS-Rest}.}
\label{fig:prompt}
\end{figure}

We employ ChatGPT\footnote{
Available at \url{https://chat.openai.com/}. The specific model used is \texttt{gpt-4-1106-preview}.
} to replace humans for comparison data annotation. 
The prompt is depicted in Figure \ref{fig:prompt}.
We continuously refine the guidelines and demonstrations in the prompt based on the annotation results.
Furthermore, to further enhance annotation quality, we adopt three strategies:
\begin{itemize}
    \item \textbf{Self-consistency}: For each sample, we input it into ChatGPT twice. We keep the sample if the results are consistent.
    \item \textbf{Self-assessment}: We require ChatGPT to evaluate its confidence level on a scale from 1 to 5 after each judgment. We retain only those samples with a confidence level is 5.
    \item \textbf{Rationale augmentation}: We instruct ChatGPT to provide reasoning and explanation before making its judgment.
\end{itemize}
Additionally, to reduce annotation costs, we integrate four samples into one prompt and annotate them at once.

\section{Ranking Objectives for Training Scorer}
\label{sec:ranking-objectives}

We optimize the pseudo-label scorer on the annotated comparison dataset and explore three ranking-based training objectives: pointwise, pairwise, and listwise approaches.
The pointwise approach classifies positive and negative samples separately and can be formulated as follows:
\begin{align}
    {\cal L}_{\textsc{point}} =&-\log p(y_{p}|x) \nonumber \\
                               &-\sum_{y_{n}}\log (1-p(y_{n}|x)),
\end{align}
where $y_p$ denotes the positive label, and $y_n$ denotes the negative label.
The pairwise approach focuses on the relative quality of labels.
We implement two pairwise training objectives, detailed as follows:
\begin{align}
    {\cal L}_\textsc{pair1}=& -\sum_{y_n}\log \sigma\left[\beta \log \frac{p(y_p|x)}{p(y_n|x)}\right],\\
    {\cal L}_\textsc{pair2}=& \hspace{0pt}\sum_{y_n} \max(0, p(y_n|x)-p(y_p|x)), 
\end{align}
where $\beta$ is a hyper-parameter.
Lastly, the listwise approach optimizes the ranking of the entire list. We design a simple listwise training objective, as outlined in Equation \ref{eq:list}.

\subsection{Experiment Results}

\begin{table}[h]
\centering
\fontsize{8.5pt}{0.8\baselineskip}\selectfont
\begin{tabular}{l cc c cc c} 
\toprule
Objectives & \texttt{ACOS-Laptop} & \texttt{ACOS-Rest} \\
\midrule
Pointwise  & \textbf{67.93} & 77.80 \\
Pairwise1  & 67.13 & 77.00 \\
Pairwise2  & 66.87 & 77.40 \\
Listwise   & 67.74 & \textbf{78.50} \\
\bottomrule
\end{tabular}
\caption{
Comparison results of four ranking-based objectives (accuracy, \%).
}
\label{tab:ranking}
\end{table}

\noindent
Experimental results in Table \ref{tab:ranking} reveal that pointwise and listwise objectives outperform two pairwise objectives, with the listwise objective being slightly better overall. Consequently, we adopt the listwise objective as the default training objective.

\section{Analysis of Reranking}
\label{sec:analysis-of-ranking}

\begin{table}[h]
\centering
\fontsize{8.5pt}{0.8\baselineskip}\selectfont
\setlength\tabcolsep{3pt}
\begin{tabular}{l cccc c} 
\toprule
 & {Laptop} & {Rest} & {Rest15} & {Rest16} & Avg\\
\midrule
ST-CS  & 45.30 & 61.66 & 50.38 & 61.08 & 54.61\\
ST-CS \& \textsc{ReRank} & 45.86 & 62.95 & 51.74 & 63.51 & +1.41\\
\hdashline[2pt/4pt]
ST-CS \& \textsc{ReRank}$^\natural$ & 63.22 & 75.63 & 66.27 & 76.29 & +15.75 \\
\bottomrule
\end{tabular}
\caption{
Experimental results of reranking ($F_1$-score, \%). ST-CS denotes self-training with confidence- and scorer-based filtering. $\natural$ indicates the performance achieved using a perfect reranker.
}
\label{tab:reranking}
\end{table}

\noindent We apply our pseudo-label scorer as a reranker to rescore the candidate labels generated by the ASQP model, selecting the highest-scoring one as the final result. Table \ref{tab:reranking} shows that this reranking step significantly improves performance on the ASQP task, resulting in an average $F_1$ improvement of 1.41\%.
Additionally, if we consider a hypothetical perfect reranker that always selects the optimal candidate, Table \ref{tab:reranking} shows that the performance gain could reach up to 15.75\%.
This significant potential boost underscores the value of further exploring the reranking step.

\begin{table}[h]
\centering
\fontsize{8.5pt}{0.8\baselineskip}\selectfont
\begin{tabular}{l cccc} 
\toprule
 & {Laptop} & {Rest} & {Rest15} & {Rest16} \\
\midrule
\multicolumn{5}{c}{\textbf{Best Candidates}}\\
1-st  & 67.70\% & 69.37\% & 61.15\% & 65.11\%\\
2-nd  & 13.77\% & 15.44\% & 17.17\% & 16.29\%\\
3-rd  & 10.44\% & 8.64\%  & 13.48\% & 11.14\%\\
4-th  & 8.09\%  & 6.55\%  & 8.19\%  & 7.46\%\\
\midrule
\multicolumn{5}{c}{\textbf{Scorer's Preferred Choices}}\\
1-st  & 67.62\% & 75.09\% & 66.48\% & 74.08\%\\
2-nd  & 17.84\% & 14.03\% & 17.77\% & 15.34\%\\
3-rd  & 8.63\%  & 6.79\%  & 9.20\%  & 6.70\%\\
4-th  & 5.91\%  & 4.08\%  & 6.55\%  & 5.18\%\\
\bottomrule
\end{tabular}
\caption{
Proportions of the best and preferred candidate labels selected by the reranker.
}
\label{tab:rerank-prop}
\end{table}

Furthermore, we rank the four candidate labels obtained via beam search according to their confidence and then analyze the distribution of the best labels and those preferred by our scorer. As illustrated in Table \ref{tab:rerank-prop}, in fewer than 70\% of cases, the candidate label with the highest confidence is considered the best. 
In comparison, our scorer tends to favor candidates with higher confidence, highlighting areas for further improvement in this reranking step.

\end{document}